\documentclass[12pt,letter]{article}

\usepackage[british]{babel}

\usepackage[a4paper,top=2cm,bottom=2cm,left=2.5cm,right=2.5cm,marginparwidth=1.75cm]{geometry}

\usepackage[square,numbers]{natbib}
\usepackage{amsmath}
\usepackage{graphicx}
\usepackage[colorlinks=true, allcolors=blue]{hyperref}
\usepackage{hyperref}
\usepackage{orcidlink}
\usepackage[title]{appendix}
\usepackage{mathrsfs}
\usepackage{amsfonts}
\usepackage{booktabs} % For \toprule, \midrule, \botrule
\usepackage{caption}  % For \caption
\usepackage{threeparttable} % For table footnotes
\usepackage{algorithm}
\usepackage{algorithmicx}
\usepackage{algpseudocode}
\usepackage{listings}
\usepackage{enumitem}
\usepackage{chngcntr}
\usepackage{booktabs}
\usepackage{lipsum}
\usepackage{subcaption}
\usepackage{authblk}
\usepackage[T1]{fontenc}    % Font encoding
\usepackage{csquotes}       % Include csquotes
\usepackage{diagbox}
\usepackage{color,soul}

\usepackage{cleveref}
\usepackage{multirow}

\usepackage{setspace}
\usepackage{titlesec}
\titleformat{\section} % Redefine section numbering format
  {\normalfont\Large\bfseries}{\thesection.}{1em}{}
  
\usepackage{lineno} % Add the lineno package

\rightlinenumbers % Right-align line numbers

\linenumbers % Enable line numbering

\usepackage{float}   % for customizing caption position
\usepackage{caption} % for customizing caption format


\crefname{algorithm}{alg.}{algorithms}
\Crefname{algorithm}{Algorithm}{Algorithms}

\title{STONet: A neural operator for modeling solute transport in micro-cracked reservoirs}

\author[1,*]{Ehsan Haghighat}
\author[2]{Mohammad Hesan Adeli}
\author[3]{S. Mohammad Mousavi}
\author[1]{Ruben Juanes}
\affil[1]{\small Massachusetts Institute of Technology, Cambridge, MA}
\affil[2]{Sharif University of Technology, Tehran, Iran}
\affil[3]{Cornell University, Ithaca, NY}
\affil[*]{Corresponding author: \texttt{ehsanh@mit.edu}}

\date{}  % Remove date

\begin{document}
\maketitle

% \begin{abstract}
% In this work, we develop a neural operator, the Solute Transport Operator Network (STONet), to model contaminant transport in micro-cracked subsurface reservoirs.
% The model combines different networks to encode heterogeneous properties effectively. 
% By predicting the concentration rate, we are able to accurately model the transport process. 
% Numerical experiments demonstrate that our neural operator approach achieves accuracy comparable to that of the finite element method.
% The previously introduced Enriched DeepONet architecture has been revised, motivated by the architecture of the popular multi-head attention of transformers, to improve its performance without increasing the compute cost. 
% The computational efficiency of the proposed model enables rapid and accurate predictions of solute transport, facilitating the optimization of reservoir management strategies and the assessment of environmental impacts. The data and code for the paper will be published at \href{https://github.com/ehsanhaghighat/STONet}{https://github.com/ehsanhaghighat/STONet}.
% \end{abstract}

% \newpage
\begin{abstract}
In this work, we introduce a novel neural operator, the Solute Transport Operator Network (STONet), to efficiently model contaminant transport in micro-cracked porous media. STONet’s model architecture is specifically designed for this problem and uniquely integrates an enriched DeepONet structure with a transformer-based multi-head attention mechanism, enhancing performance without incurring additional computational overhead compared to existing neural operators. The model combines different networks to encode heterogeneous properties effectively and predict the rate of change of the concentration field to accurately model the transport process.  The training data is obtained using finite element (FEM) simulations by random sampling of micro-fracture distributions and applied pressure boundary conditions, which capture diverse scenarios of fracture densities, orientations, apertures, lengths, and balance of pressure-driven to density-driven flow. Our numerical experiments demonstrate that, once trained, STONet achieves accurate predictions, with relative errors typically below 1\% compared with FEM simulations while reducing runtime by approximately two orders of magnitude. This type of computational efficiency facilitates building digital twins for rapid assessment of subsurface contamination risks and optimization of environmental remediation strategies. The data and code for the paper will be published at \href{https://github.com/ehsanhaghighat/STONet}{https://github.com/ehsanhaghighat/STONet}.
\end{abstract}

\textbf{Keywords}: Machine Learning; Neural Operators; Fractured Porous Media; Solute Transport.

% \newpage

% \section*{Graphical Abstract}

% \begin{figure}[H]
%     \centering
%     \includegraphics[width=1\linewidth]{figures/graphical-abstract.png}
%     % \caption{Graphical Abstract}
%     \label{fig:graphical-abstract}
% \end{figure}

% \section*{Research Highlights}
% \begin{itemize}
% \item We present STONet, a neural operator to model subsurface transport in fractured media
% \item It is designed for density-driven flow in  heterogeneous reservoirs with full-tensor permeability
% \item The neural architecture is an extension of En-DeepONet with a multi-head attention mechanism
% \item The model, once trained with high-fidelity FEM simulations, makes accurate predictions orders of magnitude faster 
% \item Data and codes are shared publicly for reproducibility and future extensions 
% \end{itemize}

% \newpage

%-------------------------------------------
% Paper Body
%-------------------------------------------
%--- Section ---%
\section{Introduction}

The depletion of freshwater resources is a pressing global challenge, particularly in regions facing severe droughts leading to the rapid exhaustion of groundwater reserves. A significant factor contributing to water quality degradation in underground aquifers is the intrusion of seawater: the higher density of saline water facilitates its rapid dispersion and mixing within freshwater aquifers, leading to the groundwater contamination \cite{1-dentz2006variable,2-diersch2002variable}. Assessing the risk of seawater intrusion and developing mitigating strategies requires quantitative modeling of coupled flow and solute transport in porous media \cite{abarca2006optimal, 3-saeedmonir2024multiscale}. These assessments are further complicated by the common occurrence of fractures in the subsurface, which can significantly alter the flow: typically, fractures exhibit higher permeability than the surrounding domain, thus profoundly modulating groundwater flow and transport \cite{juanes2002general, molinero2002numerical, 4-sebben2015seawater,5-diersch2013feflow,6-faulkner2010review, hosseini2020numerical}. Other factors that can affect the solute transport problem include, but not limited to, transport under partially saturated conditions \cite{jimenez2020impact, zhuang2021unsaturated, saeibehrouzi2024solute} and as well as other environmental and mechanical conditions such as erosion \cite{quaife2018boundary, moore2023fluid, saeibehrouzi2025non}. 

Accounting for fractures in the modeling process generally increases the complexity of the computational models of groundwater flow and transport \cite{5-diersch2013feflow, khoei2024computational}. 
However, in cases where the size of fractures is much smaller than other dimensions of interest, upscaling approaches like the equivalent continuum model can be employed to implicitly incorporate the impact of these so-called micro-fractures in the modeling framework \cite{8-oda1986equivalent,9-zhou2008flow,10-khoei2016numerical}. 
\citet{11-khoei2023modeling} employed this approach extensively in their study by introducing inhomogeneities in the form of micro- and macro-fractures into a homogeneous benchmark problem known as Schincariol \cite{12-schincariol1997instabilities,13-musuuza2009extended}. Their investigation focused on assessing the influence of micro-fractures, both in the presence and absence of macro-fractures, on solute transport in the medium. This study leverages their work to create a dataset for training a neural operator.

Numerical modeling techniques, also known as forward models, such as the finite element method (FEM) have traditionally been employed to simulate flow and transport in porous media \cite{sahimi2011flow}. Forward models rely on an accurate understanding of model parameters, which are mostly unknown for subsurface applications except at sparse observation or injection/production wells. Therefore, repeated simulations are often performed to find model parameters while matching the data at wells. Although powerful, each forward simulation is computationally intensive, particularly for realistic, three-dimensional simulations, where computational runtimes can extend to several hours or even days for a single scenario. This significant computational demand severely limits their applicability for real-time analyses and identification or optimization tasks, where numerous simulations are necessary to explore parameter spaces or identify optimal reservoir management strategies. In contrast, machine learning (ML) techniques, particularly neural operators, offer highly efficient \emph{inference} capabilities once trained, with prediction times typically reduced to seconds or less \cite{40-lu2021learning,azizzadenesheli2024neural,kovachki2024operator}. Furthermore, ML models inherently provide analytical differentiation, a feature invaluable for optimization and sensitivity analysis. As a result, ML-based surrogate models have increasingly become an attractive and practical solution for rapid assessment and optimization of complex subsurface and groundwater contamination problems.

\paragraph{Machine learning (ML)}
Over the past few years, there has been an explosive increase in the development and application of deep learning (DL) approaches, partly as a result of data availability and computing power \cite{14-lecun2015deep}. Recent advances in deep learning approaches have pushed engineers and scientists to leverage ML frameworks for solving classical engineering problems. A recent class of DL methods, namely, Phyics-Informed Neural Networks (PINNs), have received increased attention for solving forward and inverse problems and for building surrogate models with lesser data requirements \cite{15-raissi2019physics,16-karniadakis2021physics,17-cuomo2022scientific}. PINNs leverage physical principles and incorporate them into the optimization process, enabling the network to learn the underlying physics of the problem. 
The applications of this approach extend to fluid mechanics, solid mechanics, heat transfer, and flow in porous media, among others \cite{jagtap2020conservative,kharazmi2021hp,fuks2020limitations,30-haghighat2021physics,31-cai2021physics,32-niaki2021physics,33-haghighat2022physics,goswami2022physics,34-amini2022physics,xu2023physics,zhang2023physics,yan2024physics}. 
A recent architecture, namely Neural Operators, provides an efficient framework for data-driven and physics-informed surrogate modeling \cite{40-lu2021learning,41-li2020neural,42-wang2021learning,42-wang2021learning,43-wen2022u,goswami2023physics,44-kovachki2023neural,45-haghighat2024deeponet}.
Neural Operators are a class of neural networks that operate on functions rather than vectors, enabling them to capture the relationships between input and output functions. By leveraging the power of Neural Operators, it is possible to construct surrogate models that are both data-efficient and accurate. This makes Neural Operators well-suited for problems where experimental data is limited or computationally expensive to obtain. Once trained, neural operators can be used to perform inference efficiently. Neural operators have recently been used to model transport in porous media \cite{wen2023real,kim2023real,han2024surrogate}.

\paragraph{Our contributions}
In this study, we developed a neural operator specifically designed for modeling density-driven flow in fractured porous media. 
The neural operator leverages the power of deep learning to capture the complex relationships between the equivalent permeability tensor, which is a result of variations in fracture orientation and fracture density, and the pressure gradient, and outputs the spatio-temporally varying concentration field. The new architecture, as detailed in \cref{sec:stonet}, revises the previously introduced En-DeepONet \cite{45-haghighat2024deeponet} to achieve higher accuracy at the same computational cost.

A key contribution of STONet is its careful selection and encoding of input and output features tailored for the flow and solute transport problem. The input features include the equivalent permeability tensor which encodes both intrinsic permeability as well as fracture statistics (orientation, density, length, aperture), and pressure boundary conditions. The output is the rate of change of concentration, which enables flexible and stable auto-regressive prediction of the concentration field over time. Additionally, STONet introduces a novel attention block with residual connections that operates on the encoded features from the branch and trunk networks. 
% This attention mechanism allows the network to dynamically focus on the most relevant encoded features—such as those associated with fracture-induced heterogeneity and pressure gradients—at each prediction step.
The residual connections ensure stable training and facilitate the propagation of important information from the branch network, which encodes the spatially varying physical properties, throughout the network. This design leads to improved accuracy and generalization compared to other DeepONet architectures.

We generated a dataset obtained using high-fidelity finite element simulations to train and validate the neural operator. This dataset serves as a benchmark for evaluating the performance of the proposed neural operator. We applied the developed framework to predict flow patterns in porous media under heterogeneous conditions, demonstrating its ability to handle complex geological scenarios.

%--- Section ---%
\section{Governing Equations}\label{sec2}

The governing equations describing the solute transport in fractured porous media include the mass conservation equation for the fluid phase and solute component. The fluid mass conservation is expressed as 
\begin{equation}\label{continuity}
\frac{\partial}{\partial t}(\phi\rho)+\nabla\cdot(\rho \mathbf{v}_m) = 0,
\end{equation}
where $\phi$ represents the matrix porosity, $\rho$ denotes the fluid density that varies with the solute mass fraction (concentration), and $\mathbf{v}_m$ is the fluid phase velocity vector within the matrix. This velocity can be expressed in terms of pressure by applying Darcy's law as
\begin{equation}
\mathbf{v}_m = -\frac{\mathbf{k}_m}{\mu} \left(\nabla p - \rho \mathbf{g}\right).
\end{equation}
Here, $\mu$ denotes the fluid viscosity, $\mathbf{g}$ represents the gravitational acceleration, and $\mathbf{k}_m$ stands for the permeability tensor of the matrix.
In \cref{continuity}, it is assumed that the fluid is incompressible and the matrix porosity remains constant over time. 
Substituting Darcy's law into \cref{continuity} and neglecting density variations except in the terms involving gravity (Boussinesq approximation  \cite{zeytounian2003joseph, 5-diersch2013feflow}), one obtains
\begin{equation}\label{continuity2}
\nabla\cdot\left(-\frac{\mathbf{k}_m}{\mu} \left(\nabla p - \rho \mathbf{g}\right)\right) = 0.
\end{equation}
As stated earlier, the density is a function of the solute mass fraction. Assuming a linear state equation \cite{2-diersch2002variable}, the density function is expressed as
\begin{equation}\label{density}
\rho(c)=\rho_{0} + \frac{\rho_{s}-\rho_{0}}{\rho_{0}}c,
\end{equation}
where $c$ is the mass fraction, taking a value between $0$ and $1$, and $\rho_0$ and $\rho_s$ are the reference values for the fluid density at zero mass fraction (pure water) and at unit mass fraction (pure solute), respectively. 
Additionally, $\mathbf{k}_m$ is influenced by both the intrinsic permeability of the pore structure and the geometric characteristics of micro-fractures. To compute $\mathbf{k}_m$, the domain must be divided into Representative Elementary Volumes (REV). For each REV, the specific permeability matrix can be determined using the equivalent continuum model, outlined in \cite{11-khoei2023modeling}, as
\begin{equation}\label{eqs:perm_equiv}
\mathbf{k}_m = \mathbf{k}_r + \frac{1}{12\Sigma} \sum_i {b_i^3 l_i \mathbf{M}_i}.
\end{equation}
The first part of the equivalent permeability tensor \cref{eqs:perm_equiv}, i.e.,  $\mathbf{k}_r$, is an isotropic tensor related to the intrinsic permeability of the pore structure. 
Variables $b_i$ and $l_i$ correspond to the aperture and length of micro-fractures, respectively. $\Sigma$ is the volume of the REV, and $\mathbf{M}_i$ denotes the conversion matrix defined as
\begin{equation}\label{eq6}
\mathbf{M}_i = \mathbf{I} - \mathbf{n}_i \otimes \mathbf{n}_i,
\end{equation}
where $\mathbf{n}_i$ is the unit vector normal to the $i^\text{th}$ micro-fracture. Note that, while $\mathbf{k}_r$ is an isotropic tensor, $\mathbf{M}_i$ is anisotropic due to the varied orientations of the micro-fractures.

The next governing equation pertains to the conservation of mass for the solute within the fluid phase and can be written as
\begin{equation}\label{transport}
\phi \frac{\partial}{\partial t} (\rho c) + \nabla \cdot (\rho c \mathbf{v}_m) - \nabla \cdot \left(\rho \mathbf{D}_m \nabla c\right) = 0,
\end{equation}
in which Fick's law is used for the dispersive and diffusive flux of solute components.  $\mathbf{D}_m$ is the dispersion-diffusion tensor, which is a function of the velocity $\mathbf{v}_m$, and expressed as
\begin{equation}\label{Dm}
\mathbf{D}_m = \phi\tau D_m \mathbf{I} + (\alpha_L-\alpha_T)\frac{\mathbf{v}_m \otimes \mathbf{v}_m}{\mid \mathbf{v}_m \mid} + \alpha_T \mid \mathbf{v}_m \mid \mathbf{I}.
\end{equation}
Here, the first term in eq. \eqref{transport} denotes the rate of change in the mass of the solute component, while the second and third terms denote the advective and dispersive transport mechanisms of the solute component, respectively.
$D_m$ refers to the molecular diffusion coefficient associated with the matrix, $\tau$ represents the tortuosity of the porous medium, $\mathbf{I}$ is the identity tensor, and $\alpha_L$ and $\alpha_T$ denote the longitudinal and transverse dispersivities, respectively.
The solution to the aforementioned equations can be achieved through the finite element method, as detailed in \cite{11-khoei2023modeling}.

\section{STONet: Neural Operator for solute transport in fractured porous media}\label{sec:3}

Neural operators are a class of machine learning models designed to learn mappings between infinite-dimensional function spaces. They are particularly well-suited for solving parameterized partial differential equations (PDEs) that arise in various physical phenomena. Unlike traditional neural networks that operate on fixed-dimensional vectors, neural operators can handle functions as inputs and outputs, making them ideal for constructing surrogate models for continuous space--time problems.
In this section, we review neural operators in general and then provide details about the specifics of STONet. 

\subsection{Enriched DeepONet}\label{sec:endeeponet}

The goal of a neural operator is to learn a mapping $G: U \rightarrow V$ between two function spaces $U$ and $V$. For the problem of solute transport, $U$ might represent the space of initial and boundary conditions, fracture properties such as orientation, length, and opening, and medium parameters such as permeability and porosity, while $V$ represents the space of the solute concentration field over time. 
A neural operator typically consists of two main components: (1)~a feature-encoding network, known as the branch network (B); and (2)~a query network, known as the trunk network (T).
They process the input function $u \in U$ and encode relevant features.
The final output of the neural operator is obtained by combining the outputs of the branch and trunk networks through a suitable operation, such as elementwise multiplication or a learned fusion mechanism (a final network).

\begin{figure}[H]
    \centering
    \includegraphics[width=1\linewidth]{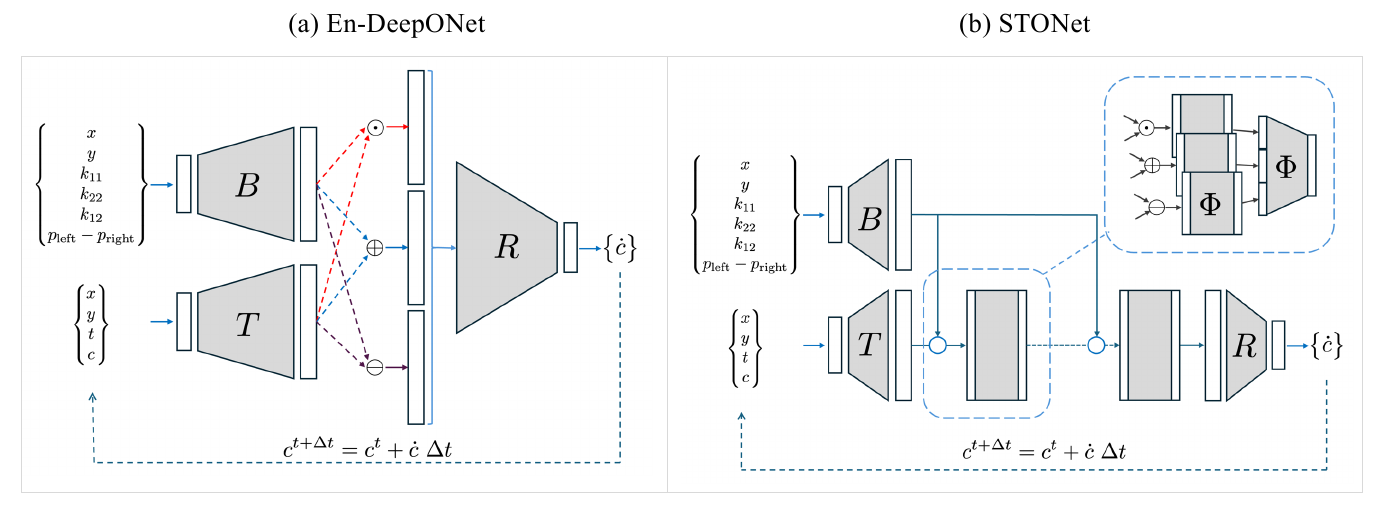}
    \caption{Network architecture. (a) En-DeepONet neural architecture \cite{45-haghighat2024deeponet}. (b) The revised En-DeepONet architecture, namely STONet. STONet resembles the multi-head attention mechanism of transformer architecture \cite{vaswani2017attention} with residual connections \cite{he2016deep} and applies the multiplication, addition, and subtraction operations on different layers.}
    \label{fig:arch}
\end{figure}

DeepONet and its generalization Enriched-DeepONet \cite{45-haghighat2024deeponet} have proven a good candidate for learning continuous functional spaces. The neural architecture for En-DeepONet is depicted in \cref{fig:arch}(a), and expressed mathematically as
\begin{align}
\epsilon_B &= B(u; \boldsymbol{\theta}_B), \\
\epsilon_T &= T(x; \boldsymbol{\theta}_T), \\
G(u)(x) &= R(\epsilon_{B \odot T}, \epsilon_{B \oplus T}, \epsilon_{B \ominus T}; \boldsymbol{\theta}_R),
\end{align}
where $x \in X$ and $u \in U$ are a query point in the solution domain $\Omega$ and a problem parameter set, respectively. 
Here, $\boldsymbol{\theta}_\alpha$ represents the set of parameters of each network, and $\epsilon_B, \epsilon_T$ are encoded outputs of the branch and trunk networks, respectively. $\odot, \oplus, \ominus$ denote elementwise multiplication, addition, and subtraction of branch and trunk encodings, respectively. $R$ is a fusion network, known as the root network, which decodes the final outputs.

\subsection{STONet}\label{sec:stonet}

Here, we present an extension of the En-DeepONet architecture that resembles the multi-head attention mechanism of transformer networks \cite{vaswani2017attention} with residual connections \cite{he2016deep}. The network architecture is depicted in \cref{fig:arch}(b). The architecture consists of an encoding branch and a trunk network. The output of the branch network is then combined with the output of the trunk network, using elementwise operations such as multiplication or addition, and passed to the new attention block. One may add multiple attention blocks. The final output is then passed on to the output root network. The network architecture is expressed mathematically as 
\begin{align}
\epsilon_B &= B(u; \boldsymbol{\theta}_B), \\
\epsilon_T &= T(x; \boldsymbol{\theta}_T), \\
\epsilon_Z^0 &= \epsilon_T, \\
\epsilon_{B \odot Z}^{l} &= \Phi(\epsilon_{B \odot Z}^{l-1}; \boldsymbol{\theta}_Z^{\odot, l}), \quad l=1 \dots L, \\
\epsilon_{B \oplus Z}^{l} &= \Phi(\epsilon_{B \oplus Z}^{l-1}; \boldsymbol{\theta}_Z^{\oplus, l}), \quad l=1 \dots L, \\
\epsilon_{B \ominus Z}^{l} &= \Phi(\epsilon_{B \ominus Z}^{l-1}; \boldsymbol{\theta}_Z^{\ominus, l}), \quad l=1 \dots L, \\
\epsilon_Z^{l} &= \Phi(\epsilon_{B \odot Z}^{l}, \epsilon_{B \oplus Z}^{l}, \epsilon_{B \ominus Z}^{l}; \boldsymbol{\theta}_Z^{l}), \quad l=1 \dots L, \\
G(u)(x) &= R(\epsilon_Z^L; \boldsymbol{\theta}_R),
\end{align}
where $\Phi$ denotes a single fully-connected layer, and $L$ is the total number of attention blocks. Lastly, the network is trained on the concentration rate, therefore concentration field is predicted auto-regressively using the forward Euler update as 
\begin{equation}
c^{t+\Delta t} = c^t + G(u)(x) \Delta t.
\end{equation}

Regarding the choice of input and output features, we performed many experiments to find the best-performing architecture, but we avoided reporting all cases here. Those experiments include:
\begin{itemize}
\item Predicting $c$ instead of $\dot{c}$. This is not a good choice because it imposes a fixed-time stepping on the auto-regressive updates, therefore less generalizable. However, we still tested the architecture but we did not observe any improvements on this dataset. 
\item Adding velocity field $\mathbf{v}_m$ as input and output features. Based on the transport \cref{transport}, it is clear $\mathbf{v}_m$ plays an important role in the transport of the contaminant in the domain. Adding this, however, did not improve the results. Note that $\mathbf{v}_m$ is strongly correlated with other features already included, including, equivalent permeability field and boundary condition ($p_\text{left} - p_{\text{right}}$).
We suspect that the reason it did not improve the results here is because of the simplicity of the domain and the overall distribution of data. However, for more generic cases, the velocity field (or pressure gradient) should also be added as input and output features, possibly as a separate network, to achieve more generalization. Worth noting that adding $\mathbf{v}_m$  increased the training time and memory requirements. 
\end{itemize}

\subsection{Optimization}
The optimization of neural operators involves adjusting the parameters of the branch, trunk, and root networks to minimize a loss function that measures the discrepancy between the predicted and true outputs. 
In this study, the network output is the concentration rate $\dot{c}$, as mentioned previously. Hence,  the loss function is based on the mean squared error between the predicted and observed solute concentrations. 
The optimization is performed using the Adam optimizer \cite{kingma2014adam}, a variant of the stochastic gradient descent. The gradients are computed using backpropagation through the computational graph of the neural operator. 
Regularization techniques, such as weight decay or dropout, may be employed to prevent overfitting and improve the generalization of the model.

% \newpage
\section{Results and Discussion}
In this section, we present the results of our numerical experiments on the performance of the proposed neural operator for surrogate modeling of solute transport in micro-cracked reservoirs. We compare the accuracy and computational efficiency of our approach to the finite element method. Additionally, we evaluate the effectiveness of the neural operator in handling different input scenarios and encoding heterogeneous properties of the porous medium. Finally, we discuss the potential applications of our model in environmental impact assessment and groundwater management.

\subsection{Problem Description and Parameter Specification}

\begin{figure}[H]
    \centering
    \includegraphics[width=1\linewidth]{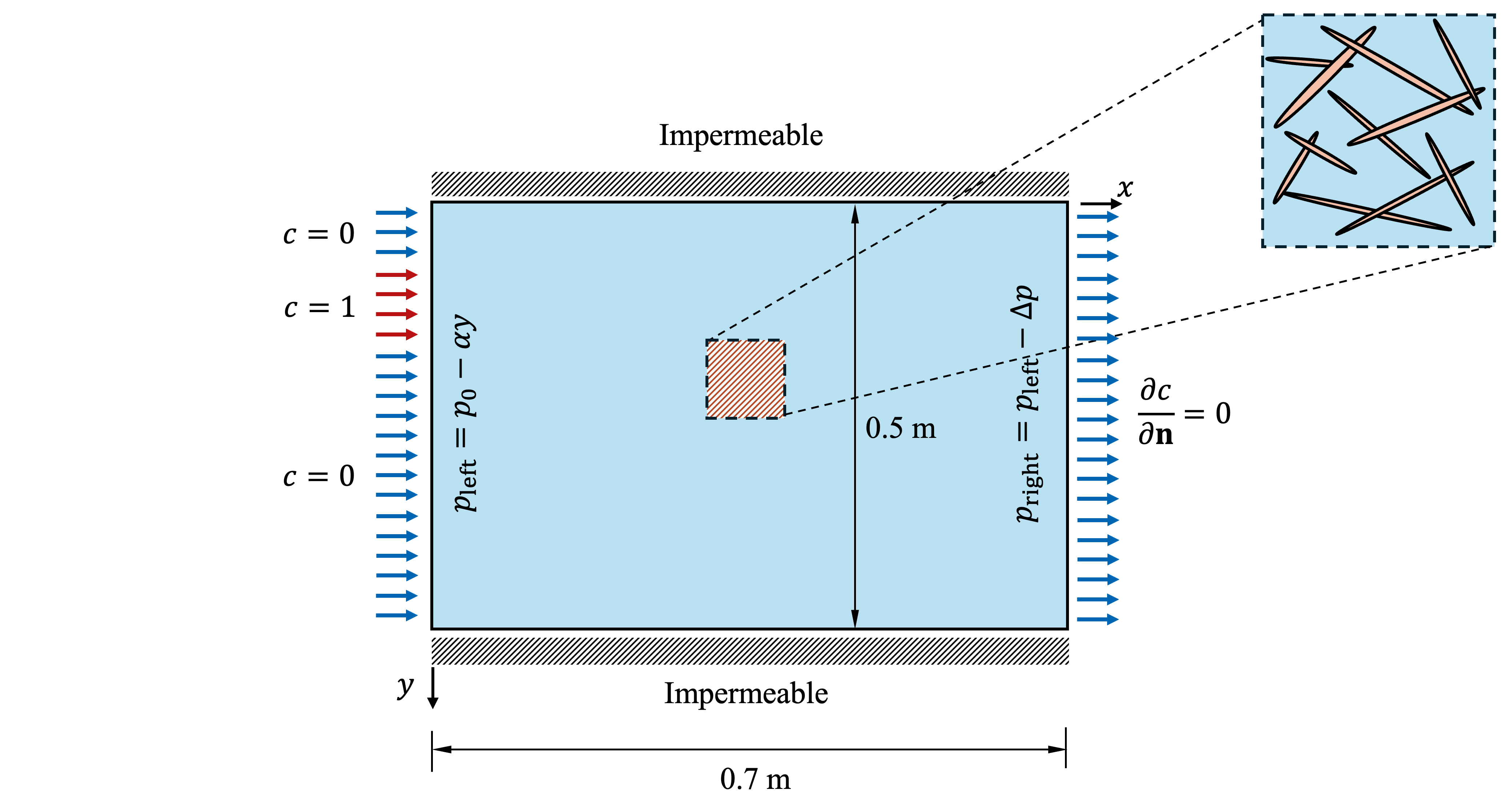}
    \caption{Problem setup. The figure depicts the original problem studied by \cite{12-schincariol1997instabilities}. The right pressure, as well as microcrack orientation and density (number of fractures per REV), are assigned randomly. }
    \label{fig:problem-setup}
\end{figure}

\begin{figure}[H]
    \centering
    \includegraphics[width=0.8\linewidth]{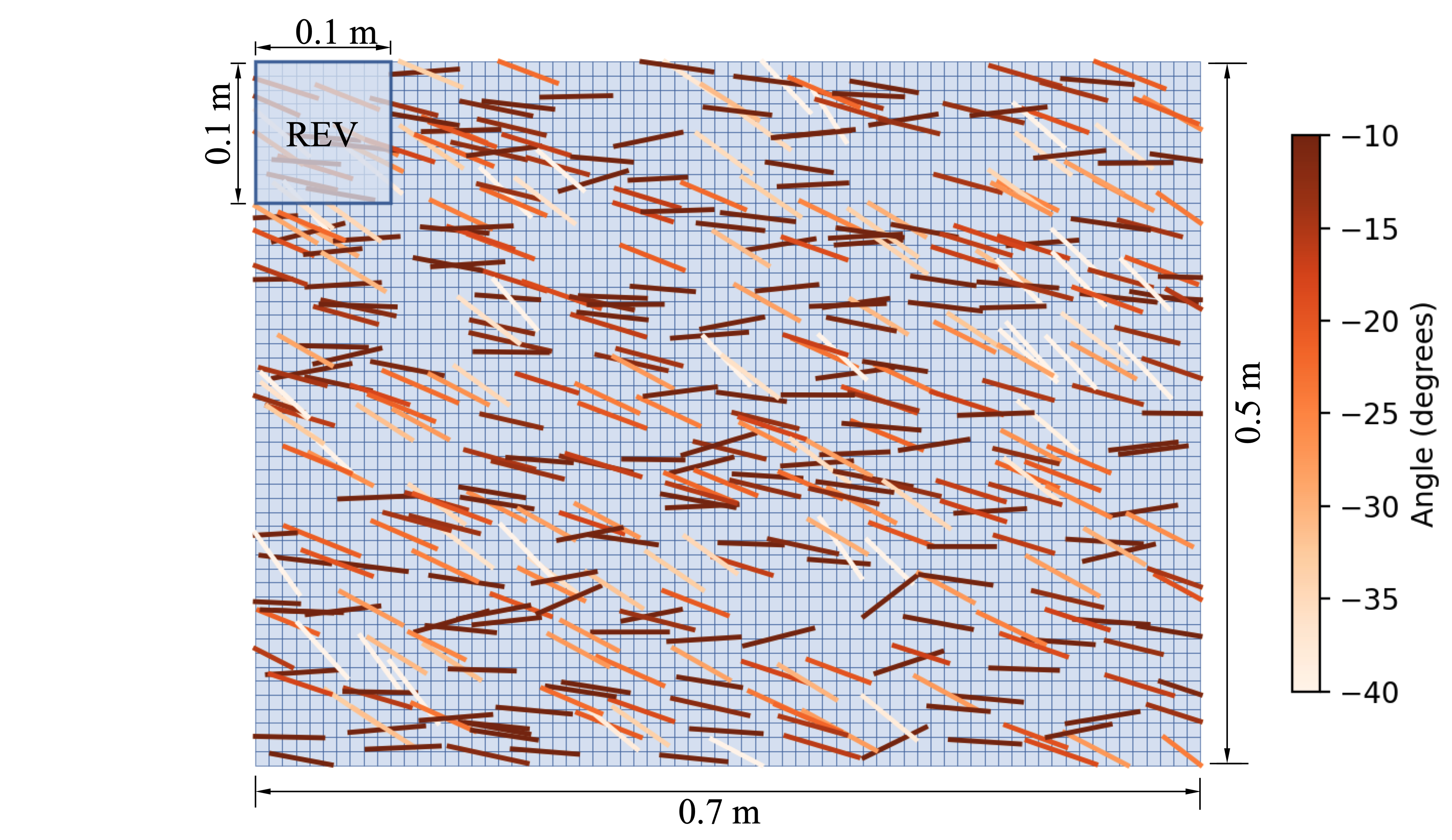}
    \caption{A sample realization of micro-crack distributions.}
    \label{fig:microcrack-distribution}
\end{figure}

A schematic of the problem considered in this study of a solute through a micro-cracked reservoir is shown in \cref{fig:problem-setup}. The domain under consideration has dimensions of $70 \times 50$~cm. The deterministic parameters of the problem are given in \Cref{tab1}. The solution is injected from the left side and transported through the domain, driven by a pressure difference between the left and right boundaries.
The porous reservoir is assumed to be confined between two impervious layers and contains randomly distributed micro-cracks of varying density and orientations.
The micro-fracture orientations are sampled randomly from a normal distribution with varying mean values but fixed standard deviation, i.e., $\theta \sim \mathcal{N}(\mu_{\theta}, \sigma_{\theta} = 15^{\circ})$, where $\mu_\theta$ is {sampled uniformly as $\mu_{\theta} \sim \mathcal{U}(-60^\circ, 60^\circ)$. }
The representative elementary volume (REV) dimensions for assessing the permeability field (using eq. (\ref{eqs:perm_equiv})) are assumed $10 \times 10$~cm.
Fracture density, i.e., the number of cracks per REV, is sampled from a Poisson's distribution as $\Sigma \sim \mathcal{P}(\lambda)$, where $\lambda$ is {sampled uniformly as $\lambda \sim \mathcal{U}(30, 70)$.} Fracture length and aperture are sampled from log-normal distributions $\log(l_c) \sim \mathcal{N}(\mu_{l_c}=0.05, \sigma_{\l_c}=0.0575$ and $\log(a) \sim \mathcal{N}(\mu_{a}=1.14\times10^{-4}, \sigma_{a}=1.15\times 10^{-4}\times 1.5)$, respectively. 
The pressure on the left side is kept fixed, while the pressure on the right side is also randomly perturbed from a normal distribution, i.e., $p_{right} = \mathcal{U}(4976, 4996) + 9792.34y$. \Cref{fig:microcrack-distribution} depicts a sample medium with micro-fractures. The sampling procedure for finding stochastic parameters is described in \Cref{alg1}.

\begin{table}[H]
\caption{Deterministic parameters.}
\centering
\begin{tabular}{lllll}
\cline{1-4}
Parameter  & Symbol      & Value    & Units   &  \\ \cline{1-4}
Gravitational acceleration & g           & 9.81     & ($m/s^2$)   &  \\
Water density              & $\rho_0$ & 998.2    & (kg/$m^3$) &  \\ Brine density              & $\rho_s$          & 1002     & ($kg/m^3$) &  \\ Viscosity                  & $\mu$  & $1.002\times10^{-3}$ & (Pa.$s$)  &  \\
Porosity                   & $\phi$  & 0.38     &         &  \\
Diffusion coefficient      & $D_m$ & $1.61\times10^{-9}$  & ($m^2$/s)  &  \\
Intrinsic permeability     & $k_r$ & $5.7\times10^{-11}$  & ($m^2$)    &  \\
Longitudinal dispersivity  & $\alpha_l$ & $1\times10^{-3}$     & ($m$)     &  \\
Transverse dispersivity    & $\alpha_t$ & $2\times10^{-4}$     & ($m$)     &  \\
Tortuosity                 & $\tau$  & 1        &         &  \\ \cline{1-4}
\end{tabular}
\label{tab1}
\end{table}

\begin{algorithm}
\caption{Sampling steps for stochastic parameters}\label{alg:cap}
\begin{algorithmic}[1]
\State $N$-samples $\gets$ Total number of samples (500).
\State $N$-quadrature $\gets$ Total number of quadrature points inside the domain.
\For{$k \gets 1$ to $N$-samples}
\State \textbf{Sample global parameters (fixed throughout the domain)}
\State $\mu_{\theta} \sim \mathcal{U}(-60^\circ, 60^\circ) \gets $  Sample $\mu_{\theta}$ uniformly between $-60^\circ$ and $60^\circ$ for the distribution of micro-fractures within the domain.
\State $\lambda \sim \mathcal{U}(30, 70) \gets $ The global Poisson distribution parameter of fracture density, sampled uniformly between 30 and 70. 

\State \textbf{Sample local parameters (at each integration point)}
\For{$i \gets 1$ to $N$-quadratures}
\State $\theta^i \sim \mathcal{N}(\mu_{\theta}, \sigma_{\theta} = 15^{\circ}) \gets$ Sample $\theta$ at every integration point inside the discretized Finite Element mesh. 
\State $\Sigma^i \sim \mathcal{P}(\lambda) \gets $ Sample fracture density at each integration point using the global value for $\lambda$. 
\State $\log(l_c^i) \sim \mathcal{N}(\mu_{l_c}=0.05, \sigma_{l_c}=0.0575 \gets$ Sample fracture length per integration point. 
\State $\log(a^i) \sim \mathcal{N}(\mu_{a}=1.14\times10^{-4}, \sigma_{a}=1.15\times 10^{-4}\times 1.5) \gets$ Sample fracture aperture per integration point. 

\State \textbf{Eval equivalent permeability tensor}
\State $\Omega_i \gets$ The Set of all quadrature points within REV defined around integration point $i$
\State $\mathbf{k}_r \gets$ Intrinsic permeability of pore structure.
\State $\mathbf{M}_i \gets$ Compute conversion matrix over $\Omega_{i}$, defined in \cref{eq6}.
\State $\mathbf{k}_m \gets $ Compute equivalent permeability tensor for integration point $i$ over $\Omega_{i}$.
\EndFor
\State $c \gets$ Solve for concentration for each sample and at each time step.
\State $\dot{c} \gets $ Eval $\dot{c}$ for each sample and at each time step using backward Euler.
\State {Store data for sample $k$ and move to next simulation.}
\EndFor
\end{algorithmic}
\label{alg1}
\end{algorithm}

\begin{figure}[H]
    \centering
    \includegraphics[width=1\linewidth]{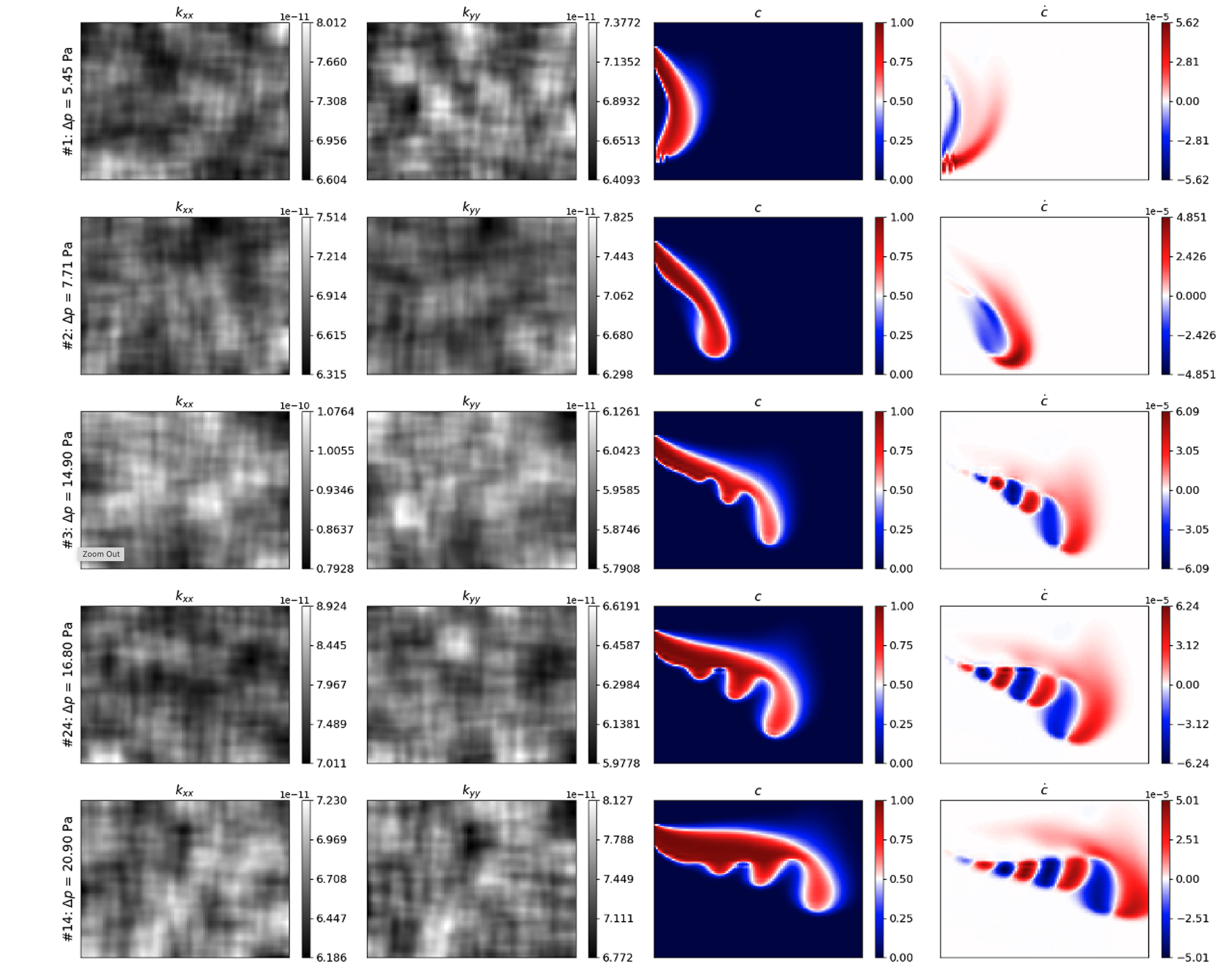}
    \caption{Sample solute concentration for different fracture patterns. Each row represents a different realization of fracture and pressure. The left two columns depict the $k_{xx}$ and $k_{yy}$ components of the equivalent permeability tensor. The right two columns represent the concentration and its rate of change, respectively, at the last time step (i.e., $t=36$~h).}
    \label{fig:sol_patterns}
\end{figure}

\begin{figure}[H]
  \centering
  \includegraphics[width=1\linewidth]{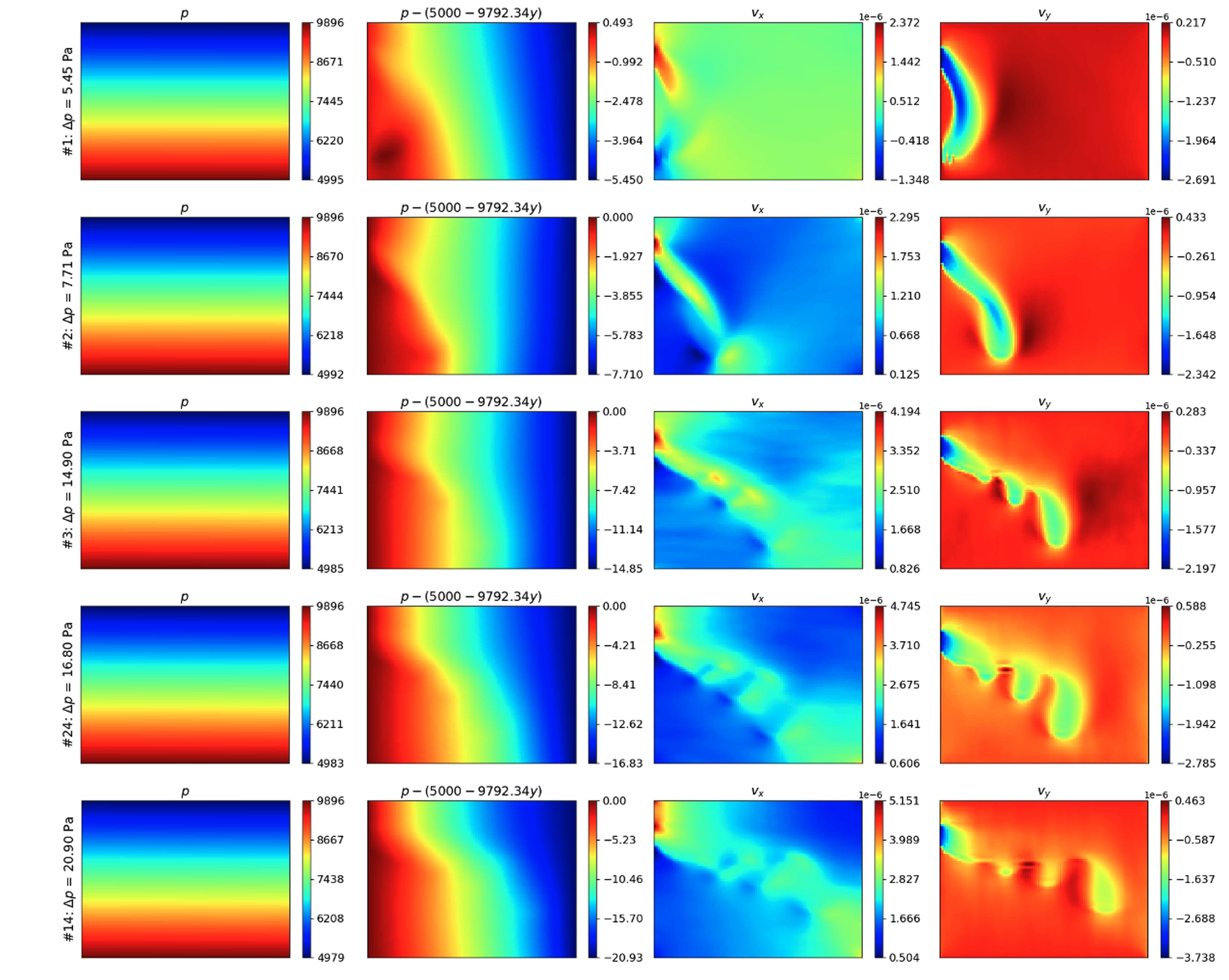}
  \caption{Sample pressure and velocity field for different fracture patterns. Each row represents a different realization, corresponding to each sample in \cref{fig:sol_patterns}. The first column shows the absolute pressure field, the second column shows the relative pressure field (by subtracting the background pressure), and the third and fourth columns show the $x$ and $y$ components of the velocity field $\mathbf{v}_m$, respectively, at the last time step (i.e., $t=36$~h).}
  \label{fig:prs_patterns}
\end{figure}

\Cref{fig:sol_patterns} depicts a few snapshots of the solute concentration for different micro-crack distributions. The equivalent permeability field $k_x$ ($k_{11}$) and $k_y$ ($k_{22}$) are plotted in the left two columns, while the right two columns highlight concentration and the rate of change of concentration at the {last time step (i.e., $t=36$ hr)}. The changes in the concentration patterns are due to different fracture orientations and pressure gradients (or velocity field $\mathbf{v}_m$), as shown in \cref{fig:prs_patterns}.

The training dataset consists of {500} FEM simulations, and the test dataset consists of 25 random unseen samples. 
The domain is discretized using an element dimension of $1\times1$~cm, resulting in a total of 3,500 elements and 3,621 nodes. We run the simulation for a total of $36$~h using $1200$~s implicit time increments. However, the outputs are recorded only at $4$~h time increments.
The generated dataset covers a wide range of fracture densities, orientations, and lengths, providing a diverse set of training examples for our neural operator.

\subsection{Sampling Strategy}

Since concentration remains near zero for a large portion of the domain (as shown in \cref{fig:sol_patterns}), to reduce the batch size and computational demand, we sub-sample 1,500 random nodes using an importance sampling strategy. Out of these, 1,000 points were selected based on concentration density, while the remaining 500 nodes were uniformly distributed over the domain. Therefore, sampling points are random in space and over time.

\subsection{Training Performance}
\begin{figure}[H]
  \centering
  \includegraphics[width=0.8\linewidth]{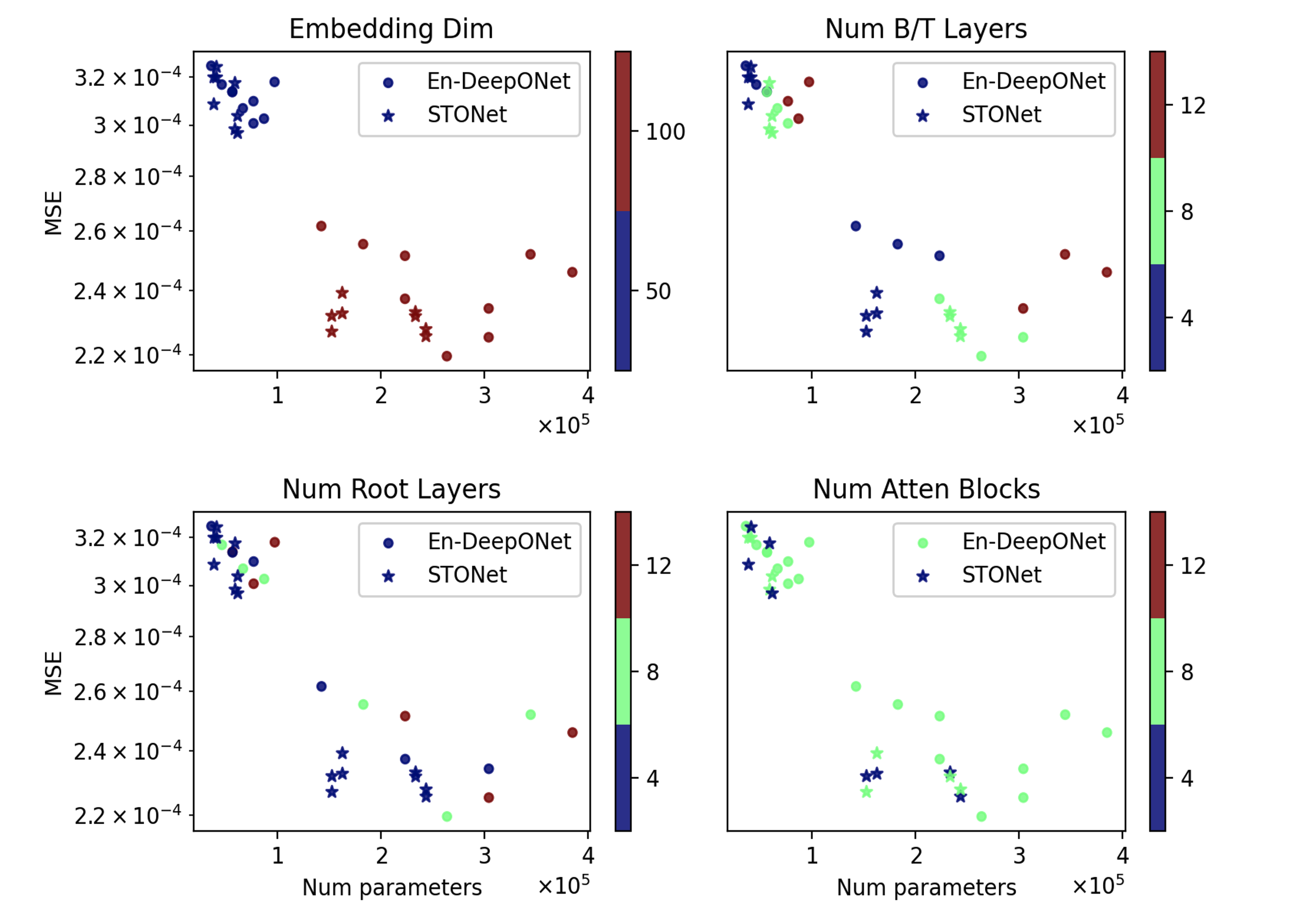}
  \caption{The training performance of the new STONet architecture versus the old En-DeepONet architecture. The x-axis shows the total number of parameters, and the y-axis presents the average loss values of the last 20 epochs for training performed for 500 epochs. Each subplot show the variations with respect to the width of the network, number of branch and trunk layers, number of root layers, and number of attention blocks, respectively. }
  \label{fig:old_vs_new}
\end{figure}

Let us first compare the performance of the new En-DeepONet architecture (i.e., STONet) with respect to the original architecture. To this end, we vary the network width and embedding dimensions in $\{50, 100\}$, the number of layers of the branch and trunk networks in $\{4, 8, 12\}$, and the number of the layers of the root network in $\{4, 8, 12\}$ and number of attention blocks in $\{4, 8\}$. 
The results are shown in \cref{fig:old_vs_new}, where circles indicate the loss with the traditional En-DeepONet architecture, and the star symbols the loss with the new STONet architecture.  We observe that for a similar number of parameters, the new STONet architecture outperforms the previous architecture without an increase in the computational cost. We also observe that both architectures improve their performance for larger network widths.

Next, to arrive at the optimal neural architecture for modeling this dataset, we explored several network sizes.
The first variable is the network width of all networks (i.e., B, T, R, $\Phi$) along with their output dimension (embedding) from $\{50, 100, 150, 200\}$. 
The second variable is the number of layers in the branch and trunk networks from $\{2, 4, 8, 12\}$. The third variable is the number of attention blocks from $\{2, 4, 8, 12\}$. The training is performed for 2,000 epochs, and the average loss for the last 100 epochs is compared. 

\begin{figure}[H]
    \centering
    \includegraphics[width=1\linewidth]{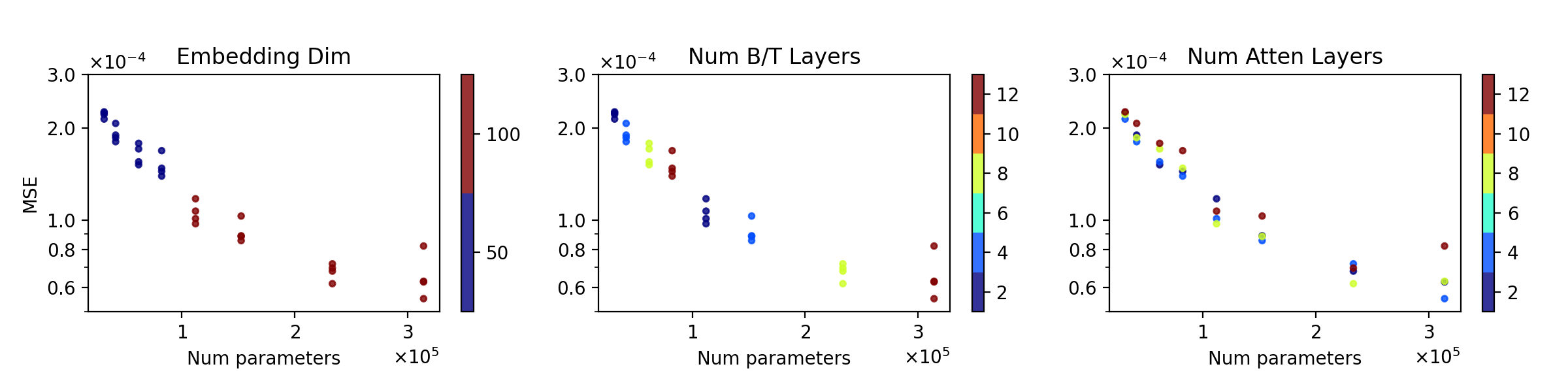}
    \caption{STONet hyper-parameter optimization. The x-axis shows the total number of parameters, and the y-axis presents the average loss values of the last 100 epochs for training performed for 2,000 epochs. }
    \label{fig:loss}
\end{figure}

The results of hyper-parameter exploration are shown in \cref{fig:loss}. The horizontal axis shows the total number of parameters, while the vertical axis presents the average of the loss value for the last 100 epochs. It is apparent that wider networks lead to a significant improvement in performance. The optimal choice of parameters seems to be 100 for network width, 8 layers for the branch and trunk networks, 8 attention blocks, and 2 root layers. Therefore, this architecture is utilized with further training (up to 50,000 epochs) to arrive at the results presented in the next section.

\subsection{Model Performance}

\begin{figure}[t]
    \centering
    \includegraphics[width=1\linewidth]{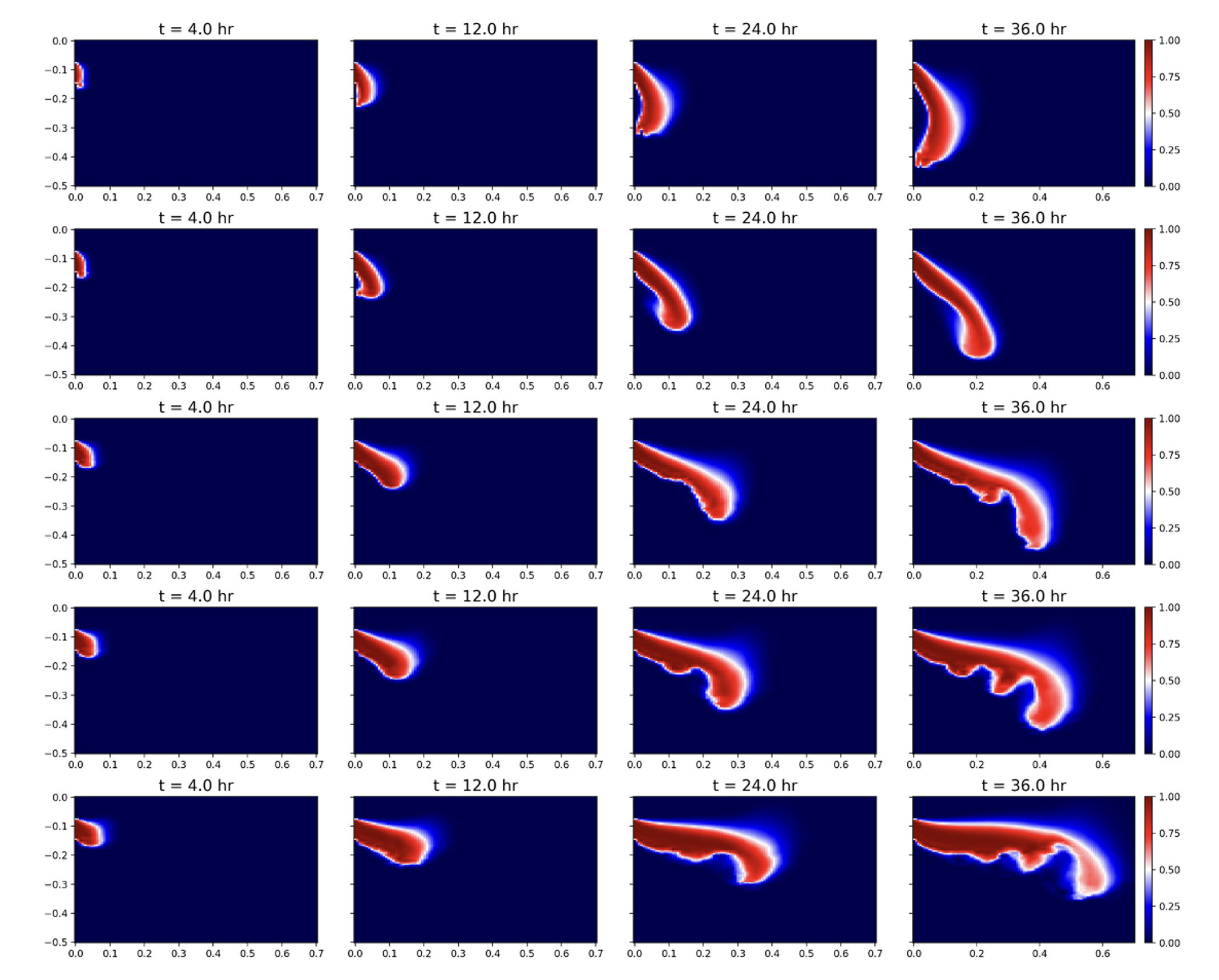}
    \caption{STONet predictions of the concentration field for five samples from the test (unseen) dataset, corresponding to each sample in \cref{fig:sol_patterns}.}
    \label{fig:test-c-preds}
\end{figure}

\begin{figure}[t]
    \centering
    \includegraphics[width=1\linewidth]{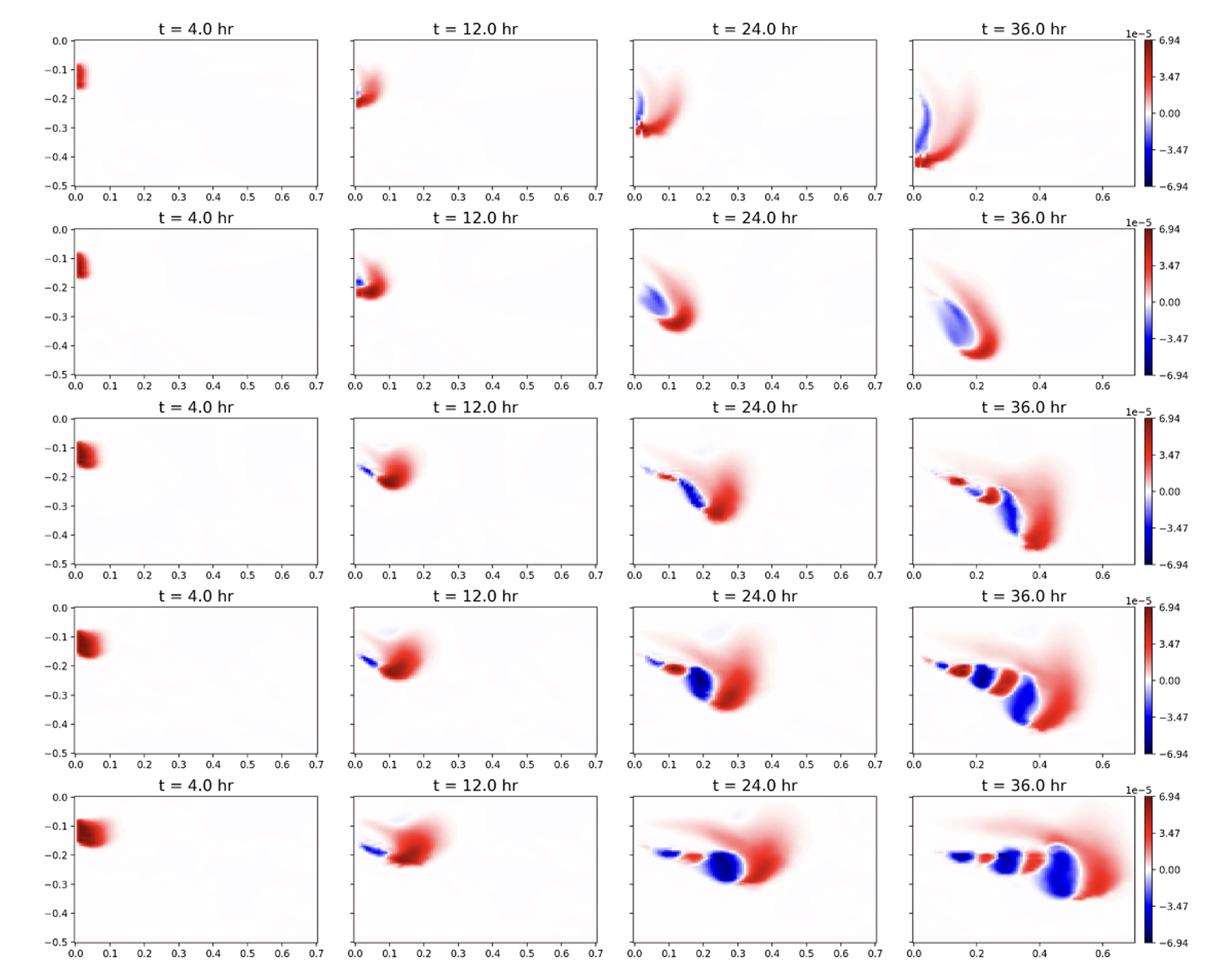}
    \caption{STONet predictions of the rate of change of concentration for five samples from the test (unseen) dataset, corresponding to each sample in \cref{fig:sol_patterns}.}
    \label{fig:test-cdot-preds}
\end{figure}

\Cref{fig:test-c-preds} depicts the predictions for the concentration on five random realizations from the test set. The corresponding predictions for the rate of change of concentration are shown in \cref{fig:test-cdot-preds}. The results from the full-physics simulations for these five test cases were shown earlier in \cref{fig:sol_patterns}. It is apparent that STONet predicts the full-physics results accurately.
The fundamental difference is that the STONet, having been pre-trained, can be used for fast prediction of density-driven flow and transport with any new fracture network, while the FEM simulation would need to be recomputed altogether for any new configuration.

\begin{figure} %[H]
    \centering
    \includegraphics[width=1.0\linewidth]{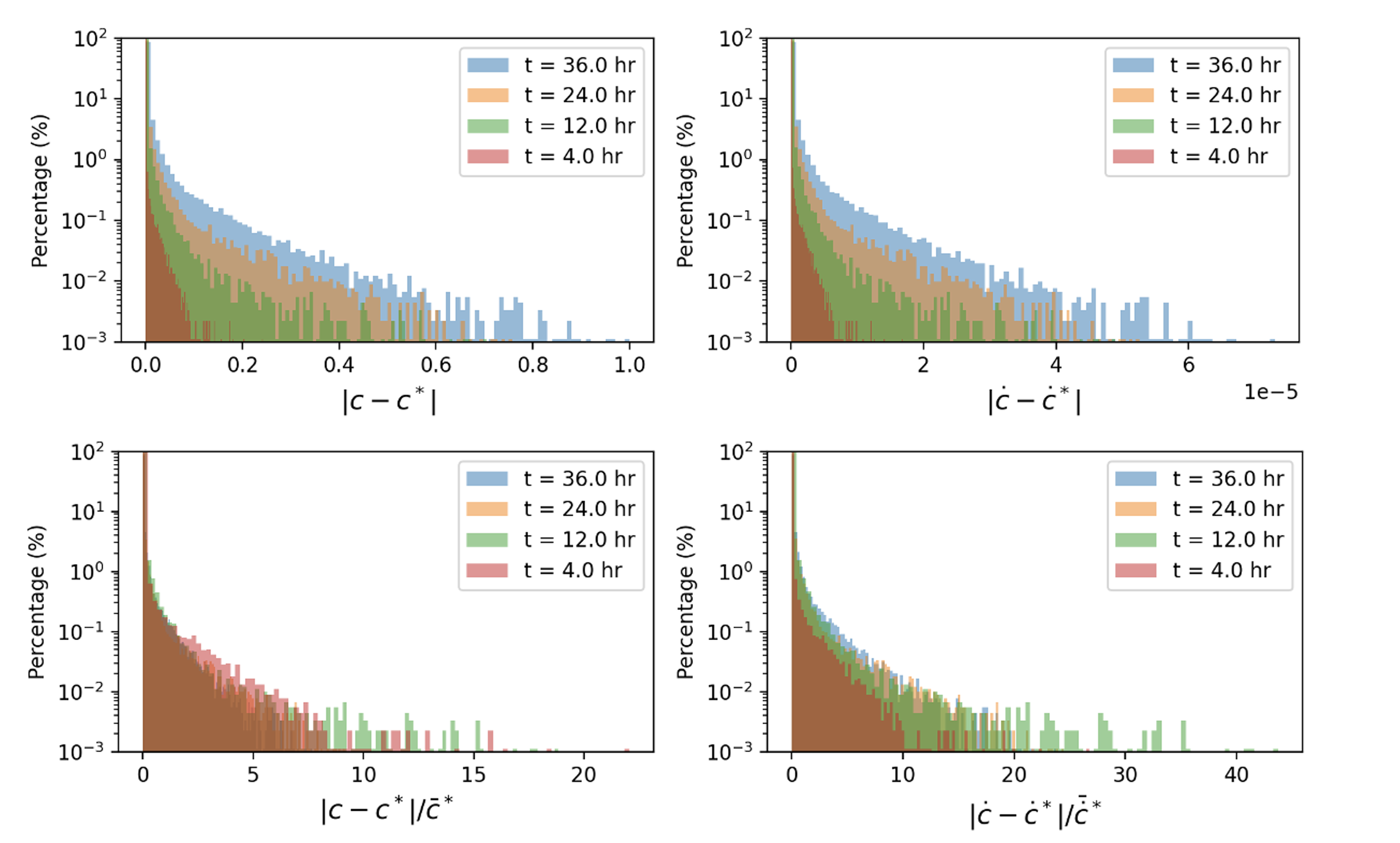}
    \caption{Unrolling error distribution. Pointwise absolute (top row) and relative (bottom row) error distribution for STONet predictions on all 25 test samples at different time steps for (Left) concentration and (Right) concentration rate. The error in the majority of points remains within 1\%. While widening absolute error distributions point to error accumulation, the distribution of relative errors remains nearly constant, which is desirable. }
    \label{fig:error-dist-time}
\end{figure}

\begin{figure} %[H]
    \centering
    \includegraphics[width=1.0\linewidth]{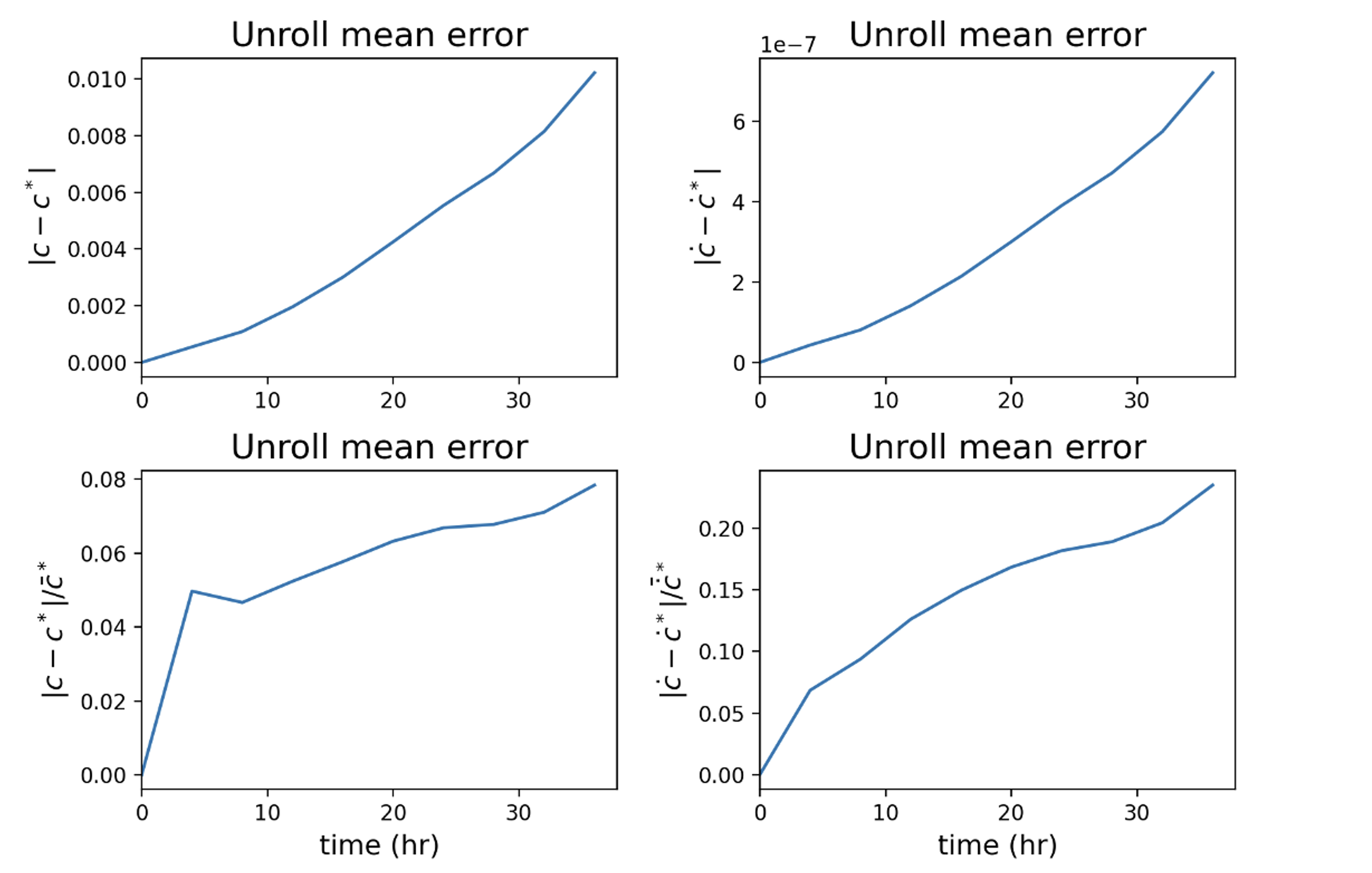}
    \caption{Unrolling error. The plot shows the mean absolute (top row) and relative (bottom row) error of the unrolling process as a result of the auto-regressive nature of the architecture. While the evolution of mean absolute error points to error accumulation, the evolution of mean relative errors remains nearly constant, which is desirable.}
    \label{fig:unroll-error}
\end{figure}

The distributions for pointwise absolute and relative error at different time steps for concentration and concentration rate are plotted in \Cref{fig:error-dist-time}. Overall, STONet's prediction error is very small, with most predictions having below 1\% error. The top figures depict the absolute error distributions and highlight the presence of accumulation error as can be observed from widening distributions at different time steps. However, the distribution of relative error at different time steps, as shown in the bottom figures, remains nearly unchanged, indicating a predictable fixed error distribution over time, which is highly desirable.
This is also confirmed by inspecting the mean absolute and relative error evolution over time, as plotted in \cref{fig:unroll-error}. It is worth noting that the accumulation error might be controlled with the addition of observational data and additional training samples using data assimilation techniques such as Active Learning \cite{wang2023inverse}.

%--- Section ---%
\section{Conclusions}
In this study, we have presented a new Enriched-DeepONet architecture, STONet, for emulating density-driven flow and solute transport in micro-cracked reservoirs.
Our approach effectively encodes heterogeneous properties and predicts the concentration rate, achieving accuracy comparable to that of the finite element method. 
The computational efficiency of STONet enables rapid and accurate predictions of solute transport, facilitating both parameter identification and groundwater management optimization. 

The STONet model developed in this work has the potential to be applied for fracture network identification and efficient tracing and control of solute transport in micro-cracked reservoirs. By rapidly and accurately predicting the concentration rate, the model can help identify the location and connectivity of fractures in the porous media, which is crucial for optimizing the management of groundwater resources. Additionally, the model can be used to predict the transport of solutes in different scenarios, such as accidental contaminant spills, allowing for accurate and efficient decision-making. Overall, the ML model has the potential to significantly improve the sustainable management of underground aquifers, contributing to both local and global efforts towards sustainable groundwater resource utilization.

While STONet demonstrates promising results for modeling solute transport in micro-cracked reservoirs, several limitations remain. The current study is restricted to a single domain geometry and accounts for limited variations in permeability but doesn't account for the presence of distinct geological layers. Additionally, the model assumes a fixed injection location for the contaminant and does not consider scenarios where the contaminant flow originates from multiple or varying locations. The study also does not account for different pressure or flux boundary conditions, which are critical for capturing diverse real-world scenarios. Furthermore, the study is limited to a two-dimensional domain, whereas real-world applications often require three-dimensional modeling to capture the full complexity of subsurface transport. Considering all these choices may require hundreds of thousands of training samples, which is out of the scope of the current study. These limitations highlight the need for further research and development to extend the applicability of STONet to more generic and realistic scenarios.

%-------------------------------------------
% Optional Contents
%-------------------------------------------

%--- Section ---%
\section*{Conflicts of Interest} 
The authors have no conflicts of interest to declare.

%--- Section ---%
\section*{Author Contributions}
EH conceptualized the problem and contributed to the theoretical development and implementation. MHA contributed to the implementation, running the test cases and generating the figures. MM prepared the training dataset. All authors contributed to drafting and revising the manuscript. 

%--- Section ---%
% \section*{Acknowledgments}
% The authors thank the anonymous reviewers for their valuable suggestions. 

\section*{Data availability}
The data and code for reproducing the results reported in this manuscript will be published at \href{https://github.com/ehsanhaghighat/STONet}{https://github.com/ehsanhaghighat/STONet}. 

%-------------------------------------------
% References
%-------------------------------------------

% Print bibliography
% \printbibliography
\bibliography{references.bib}

%-------------------------------------------
% Appendix
%-------------------------------------------
% Activate the appendix in the doc
% from here on sections are numerated with capital letters 
%\appendix

% Change equation numbering format to be sequential within sections in the appendix
\renewcommand\theequation{\Alph{section}\arabic{equation}} % Redefine equation numbering format
\counterwithin*{equation}{section} % Number equations within sections
\renewcommand\thefigure{\Alph{section}\arabic{figure}} % Redefine equation numbering format
\counterwithin*{figure}{section} % Number equations within sections
\renewcommand\thetable{\Alph{section}\arabic{table}} % Redefine equation numbering format
\counterwithin*{table}{section} % Number equations within sections

% Appendix
% \begin{appendices}
% \section{Section  name}
% \end{appendices}

\end{document}